\documentclass[conference]{IEEEtran}
\usepackage{times}

\usepackage[numbers]{natbib}
\usepackage{multicol}
\usepackage[bookmarks=true]{hyperref}
\usepackage[dvipsnames]{xcolor}
\usepackage{xspace}
\usepackage{siunitx}
\usepackage{amsmath}
\usepackage[font=small]{caption} 
\usepackage[font=small]{subcaption}
\usepackage[linesnumbered,ruled,vlined]{algorithm2e}
\usepackage{amsfonts}
\usepackage[capitalise]{cleveref}

\usepackage{caption}
\usepackage{graphicx}

\usepackage{tabularx}

\newcommand{\algname}{RLS\xspace}
\newcommand{\rc}{\emph{robot classrooms}\xspace}
\newcommand{\algfullname}{RL@Scale\xspace}

\newcommand{\E}{\mathbb{E}}

\newcommand{\act}{\mathbf{a}}
\newcommand{\s}{\mathbf{s}}

\begin{document}

\title{Deep RL at Scale: Sorting Waste in Office Buildings with a Fleet of Mobile Manipulators}

\author{
\authorblockN{Alexander Herzog\authorrefmark{1}\authorrefmark{2},
Kanishka Rao\authorrefmark{1}\authorrefmark{3},
Karol Hausman\authorrefmark{1}\authorrefmark{3},
Yao Lu\authorrefmark{1}\authorrefmark{3},
Paul Wohlhart\authorrefmark{1}\authorrefmark{2},
}
\authorblockN{Mengyuan Yan\authorrefmark{2},
Jessica Lin\authorrefmark{2},
Montserrat Gonzalez Arenas\authorrefmark{3},
Ted Xiao\authorrefmark{3},
Daniel Kappler\authorrefmark{2},
Daniel Ho\authorrefmark{2},
}
\authorblockN{Jarek Rettinghouse\authorrefmark{2}
Yevgen Chebotar\authorrefmark{3},
Kuang-Huei Lee\authorrefmark{3},
Keerthana Gopalakrishnan\authorrefmark{3},
Ryan Julian\authorrefmark{3},
Adrian Li\authorrefmark{2},
}
\authorblockN{
Chuyuan Kelly Fu\authorrefmark{2},
Bob Wei\authorrefmark{2},
Sangeetha Ramesh\authorrefmark{2},
Khem Holden\authorrefmark{3},
Kim Kleiven\authorrefmark{2},
David Rendleman\authorrefmark{3},
}
\authorblockN{
Sean Kirmani\authorrefmark{2},
Jeff Bingham\authorrefmark{2},
Jon Weisz\authorrefmark{2},
Ying Xu\authorrefmark{2},
Wenlong Lu\authorrefmark{2},
Matthew Bennice\authorrefmark{2},
Cody Fong\authorrefmark{2},
}
\authorblockN{
David Do\authorrefmark{2},
Jessica Lam\authorrefmark{2},
Yunfei Bai\authorrefmark{2},
Benjie Holson\authorrefmark{2},
Michael Quinlan\authorrefmark{2},
Noah Brown\authorrefmark{3},
}
\authorblockN{
Mrinal Kalakrishnan\authorrefmark{2},
Julian Ibarz\authorrefmark{3},
Peter Pastor\authorrefmark{2},
Sergey Levine\authorrefmark{3}
}
\authorblockA{\authorrefmark{1}Authors with equal contribution ~
\authorrefmark{2}\href{https://everydayrobots.com/}{Everyday Robots} ~
\authorrefmark{3}\href{http://g.co/robotics}{Robotics at Google}}
}

\maketitle

\begin{abstract}
We describe a system for deep reinforcement learning of robotic manipulation skills applied to a large-scale real-world task: sorting recyclables and trash in office buildings. Real-world deployment of deep RL policies requires not only effective training algorithms, but the ability to bootstrap real-world training and enable broad generalization. To this end, our system combines scalable deep RL from real-world data with bootstrapping from training in simulation, and incorporates auxiliary inputs from existing computer vision systems as a way to boost generalization to novel objects, while retaining the benefits of end-to-end training. We analyze the tradeoffs of different design decisions in our system, and present a large-scale empirical validation that includes training on real-world data gathered over the course of 24
months of experimentation, across a fleet of 23 robots in three office buildings, with a total training set of 9527 hours of robotic experience. Our final validation also consists of 4800 evaluation trials across 240 waste station configurations, in order to evaluate in detail the impact of the design decisions in our system, the scaling effects of including more real-world data, and the performance of the method on novel objects. The projects website and videos can be found at \href{http://rl-at-scale.github.io}{rl-at-scale.github.io}.
\end{abstract}

\IEEEpeerreviewmaketitle

\section{Introduction}

Real-world robotic manipulation problems require the integration of a range of components, including visual perception, planning, and control. The design and integration of these components, and the abstractions needed to make them work together, often present a challenge for real-world deployment of robotic systems in open-world settings. End-to-end learning offers an appealing alternative: by optimizing the performance of a robotic learning system directly on the final objective of the task, and learning this task directly from visual observations, we can in principle sidestep these challenges and devise robotic manipulation strategies that are fully optimized for the real-world task that they are intended to solve. However, while general-purpose reinforcement learning algorithms can in principle provide such functionality, in practice they suffer from their own set of challenges, including difficulties associated with gathering suitable datasets, generalization, and overall system design. While a broad range of robotic learning methods have been proposed, it remains unclear how to devise an end-to-end learning system that can scale to realistic real-world tasks. In this paper, we approach this system design problem in the context of a complex real-world problem that requires effective and generalizable manipulation strategies: sorting trash and recyclables in office building waste bins. This task naturally provides a broad range of objects for goal-directed manipulation, and serves as a lens to examine how we can design effective and generalizable robotic reinforcement learning pipelines for the real world. The central question in this paper is: how can the various tools in the deep reinforcement learning toolbox be put together to address such a complex and diverse real-world task?
\begin{figure}[t]
    \centering
    \includegraphics[width=0.99\linewidth]{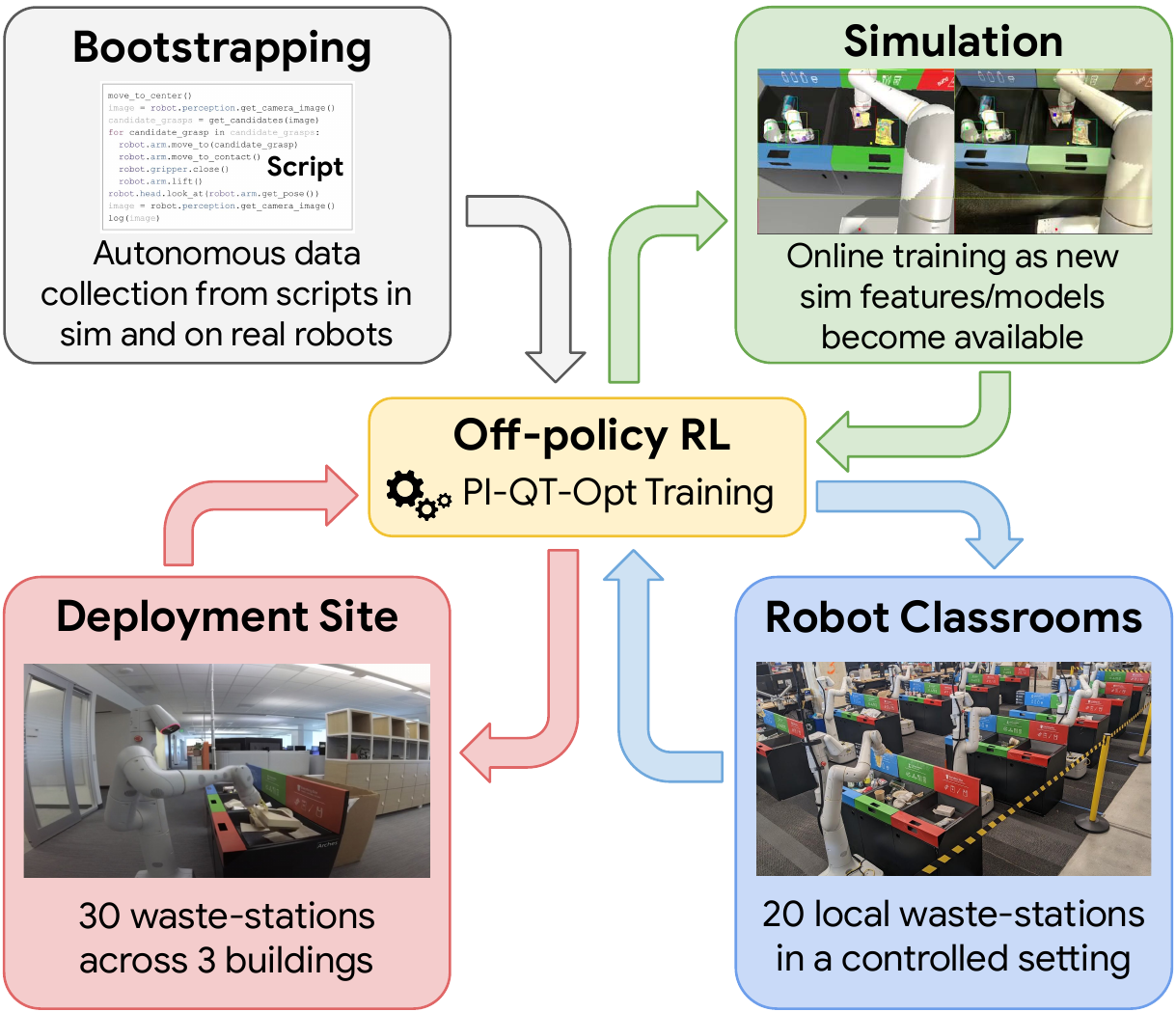}
    \caption{Overview of our data flywheels that we operated over 24~months: We bootstrap the initial policy from scripts in simulation and on real robots (grey), re-train the policy in simulation as needed (green), deploy the latest policy weekly to a local setup of 20 robots sorting 20 waste stations on random waste-scenes and scenes encountered in the deployment site (blue), and deploy to 23 robots operating in 3 different buildings sorting 30 waste stations (red).}
    \label{fig:overview}
    \vspace{-0.5cm}
\end{figure}
Prior works have proposed a variety of tools of this sort, including the use of simulation to overcome the high sample complexity requirements of reinforcement learning from scratch~\cite{OpenAI2019SolvingRC,sim2real_survey20}, the use of prior data~\cite{singh2020cog,kumar2022pre,awopt22}, off-policy or offline reinforcement learning strategies~\cite{levine2020offline}, large-scale collective learning involving fleets of multiple robots~\cite{qtopt18,mtopt21}, and the use of priors and inductive biases~\cite{muller2018driving,sax2018mid,jaderberg2016reinforcement}. However, each of these approaches has their own strengths and weaknesses. Simulated training is problematic if we expect to encounter a broad range of real-world situations, as unless we can characterize this range of settings in advance, we might fail to generalize. Collective learning from real-world data gathered by multiple deployed robots provides a very powerful method for gathering data that is representative of what the robot actually encounters in the real-world, but training from scratch in the real world presents a major exploration challenge, and relying entirely on robot-collected data can make generalization challenging. Incorporating prior knowledge from other sources, such as computer vision datasets or data from the Internet, can provide a major boost to generalization, but it remains unclear how to most effectively integrate it into an end-to-end trained system. Our final system, which we call \algfullname or \algname in short, makes use of all three components: real-world training is bootstrapped using simulation, collective learning by a fleet of 20 robots is used to collect representative experience in the real world, and a visual masking approach is further used to incorporate prior knowledge about object classes and appearance from computer vision datasets to additionally boost generalization to new objects. Our extensive empirical investigation shows the importance of each component, and provides a large-scale case study of how end-to-end deep RL can be deployed for a practical and relevant real-world task at scale.

Our evaluation requires the robot to sort trash and recyclables: the robot locates bins with recyclables, approaches them, and sorts their content by type to minimize contamination (i.e., moving all compostables, recyclable plastic, and trash into their respective containers). This task requires generalization to a wide variety of possible objects, as well as the ability to pick up and identify those objects so as to place them in the right container. The objects that the robot might encounter will include difficult-to-simulate properties, such as in the case deformable chip bags, or might be particularly hard to grasp, such as in the case of large boxes. They will also vary over time and across locations. Thus, this task presents a particularly challenging manipulation scenario, but also one that provides a great testbed for scalable robotic learning.

The contribution of our work is a system for real-world end-to-end deep reinforcement learning, \algname, for a complex robotic manipulation task. While our policies are trained end-to-end, they incorporate additional knowledge via object masks trained on computer vision datasets, and are bootstrapped from simulation to overcome the exploration challenge. We show that these decisions lead to excellent performance in the real world. While the individual components of this system have been explored in various prior works, we focus on their integration into a complete robotic learning system at scale: we believe that our method is the first to be applied at such a large scale to a realistic real-world task, and therefore serves as a valuable case study to understand the considerations and tradeoffs in the design of end-to-end robotic learning systems. We present extensive ablations and comparisons to study the individual components of our system, and show that our best-performing design can sort $84.35\%$ of misplaced objects on challenging waste scenarios. We demonstrate dexterous robotic manipulation of waste that emerge from the end-to-end formulation of our policy. When deployed at office buildings, our fleet of mobile manipulators further reduce contamination of waste stations filled by office workers unrelated to the project as part of their day-to-day by up to $53\%$.   

\section{Related Work}
\label{sec:rw}

Advances in reinforcement learning algorithms~\citep{DRLsurvey17,DRL_nav_survey21, DRLsurvey_drone21} have enabled machine learning systems that can play games~\citep{atari15,rl_selfplay18}, control robots~\citep{manipulation_review19,qtopt18, OpenAI2019SolvingRC, cabi20, mtopt21}, and perform a variety of tasks from chip design to drug discovery~\citep{mirhoseini2020chip,popova2018deep}. However, real-world deployment of RL systems, particularly in robotics, presents a number of major challenges~\citep{realworld_RL19}.

First, deep RL algorithms typically require a large number of samples, especially when learning large, image-based policies~\citep{atari15}. Because of this, a common choice is to employ simulation for training~\citep{sadeghi2016cad2rl,matas2018sim,rcan19,ho2021retinagan}. While policies trained in simulation can work well in relatively constrained real-world settings in a laboratory~\citep{matas2018sim,OpenAI2019SolvingRC,chebotar2019closing}, simulated training alone can be insufficient for robotic systems that must generalize to a wide variety of real-world environments, where the range and variability of objects and settings might exceed the variability seen in the simulator. In our proposed robotic RL system, we employ simulated data generation, but combine it with real-world data that is collected autonomously through a variety of policy bootstrapping approaches, and find that this hybrid design allows our system to generalize well in the real world.

Second, RL agents that learn complex tasks from scratch must typically spend considerable time on exploration to discover effective behaviors~\citep{bellemare2016unifying,osband2016deep,tang2017exploration}, particularly with simple sparse reward functions that merely indicate success or failure at the task and provide little guidance for how to improve before a successful behavior is discovered. To avoid this challenge, many prior works have explored imitation learning as an alternative to RL~\citep{zhang2018deep,mandlekar2021matters,shridhar2022perceiver,jang2022bc}, or as a way to supplement RL~\citep{demos_and_rl17, riedmiller18, awopt22} in robotics. In our system, we also sidestep exploration, but we find that we can do this effectively with a combination of relatively simple scripted exploration policies, bootstrapping from simulation, and the use of multi-task training to learn simple tasks as a stepping stone to more complex ones. While prior works have studied each of these methods individually~\citep{qtopt18,riedmiller18,smith2022legged}, our system combines all of these components into a complete robotic manipulation system that tackles a complex real-world task.

Third, RL-trained policies, like any machine learning model, are vulnerable to distributional shift: when the conditions at deployment-time do not match the conditions seen in training, a learned model will underperform. RL provides an appealing solution to this problem: as the agent experiences new domains, it can simply keep training and continue to adapt~\citep{finetune21julian}. While this observation is not new, we show in our work how a complete RL-based robotic manipulation system can benefit from this capability and get better as it observes more real-world data.

Our work is related to a number of large-scale learning-enabled robotic manipulation systems that have been proposed in prior work. Many of these systems have been shown to learn behaviors that are physically complex~\citep{gps17, OpenAI2019SolvingRC} or exhibit good generalization~\citep{qtopt18, riedmiller18, cabi20, mandlekar2020iris, mtopt21, awopt22, lee2022pi}, but their evaluations are typically confined to laboratory settings, or else to tasks such as locomotion or navigation that do not require manipulating a wide variety of objects~\citep{hwangbo2019learning,loquercio2019deep,kumar2021rma}. In contrast, our aim is to develop and evaluate a system that can be deployed at scale on a realistic object manipulation task, with a real-world deployment and a quantitative evaluation that evaluates on scenes replicated from this deployment.

\section{Preliminaries}
\label{sec:prelims}
Let $\mathcal{M} = (\mathcal{S}, \mathcal{A}, P, R, p_0, \gamma)$ define an MDP, where $\mathcal{S}$ and $\mathcal{A}$ are state and action spaces, $P$ is a state-transition probability function, $R$ is a reward function, $p_0$ is an initial state distribution, $\gamma$ is a discount factor. 
We use a sparse reward function that assigns a reward of $1.0$ if the task was accomplished successfully at the end of the episode, and $0.0$ otherwise.
The goal of reinforcement learning is to find a policy $\pi(\act | \s)$ that maximizes the expected discounted reward over trajectories induced by the policy,
$\E_{\pi}[R(\tau)]$, where $ \s_0\sim p_0, \s_{t+1}\sim P(\s_{t+1} | \s_t, \act_t)$ and $\act_t\sim \pi(\act_t | \s_t)$.

To train our end-to-end sorting policy, we use PI-QT-Opt~\cite{lee2022pi}, a combination of QT-Opt~\cite{qtopt18} and a predictive information representation auxiliary~\cite{lee2020predictive,lee2022piars,lee2022pi}. 

QT-Opt is a value-based RL method that learns a Q-function by minimizing the Bellman error, and obtains the policy by running a stochastic optimization on the learned Q-function using cross-entropy method (CEM)~\cite{rubinstein2004cross}, a zeroth-order optimizer that finds $\arg\max_{\act_t}Q_\theta(\s_t,\act_t)$ for the learned Q-function $Q_\theta(\s_t,\act_t)$. 
Importantly, for CEM to be effective in this setting, the actions generated by the RL agent should be in distribution of the actions sampled by the CEM in the Bellman update equation described below.
To learn the Q-function, QT-Opt optimizes the following objective:
$
    \mathcal{L}(\theta) = \E_{(\s_t,\act_t,\s_{t+1})\sim p(\s_t,\act_t,\s_{t+1})} {\big[ D (Q_\theta(\s_t,\act_t), Q_T(\s_t, \act_t, \s_{t+1})) \big]},
$
where $Q_T(\s_t,\act_t,\s_{t+1}) = r(\s_t, \act_t) + \gamma \max_{\act'} Q_\theta(\s_{t+1}, \act')$
is a target Q-value, $\gamma$ is a discount factor, and $D$ is a divergence metric. Since the cumulative reward for each episode is always between $0.0$ and $1.0$, the Q-values can be treated as probabilities (formally, denoting the probability that the robot will succeed at the task), and thus we can use the cross-entropy loss for $D$. The target value $\max_{\act'} Q_\theta(\s_{t+1}, \act')$ is computed by running the same CEM optimization as we use for the policy action selection.
In this work, we use the version of QT-Opt that includes predictive information called PI-QT-Opt~\cite{lee2022pi}. We provide the preliminaries of predictive information in the appendix.

\section{Task and System Overview}
\label{sec:overview}

Our aim is to develop a deep RL system that can be applied to a real-world waste sorting task, exhibits robust performance and good generalization, and can integrate both autonomously collected real-world experience and bootstrapping from simulation and computer vision systems. In this section, we first describe the waste sorting task we tackle, and then provide an overview of the different components of our system.

\subsection{Waste Sorting Task}
We situate our deep RL system in the context of a waste sorting task, which requires separating trash, compostables, and recyclables in waste bins.
Since office buildings are created to serve a large number of people, they usually have multiple trash stations distributed throughout different floors, resulting in broad range of environments and potential trash sorting scenarios.
Each sorting station contains three bins that correspond to recyclables, compost and landfill.
Office workers unrelated to the project use these stations in their regular day-to-day routines. Usually, they make mistakes and misplace their trash into incorrect bins resulting in bin contamination. 
Our goal is to deploy an autonomous robot to reduce the resulting contamination (i.e., move all the compostables, recyclables, and landfill waste items into their respective bins).
We present the visual depiction of the task in Fig.~\ref{fig:robot_and_task}.
This task presents a number of challenges for a robotic manipulation system: (i) the robots must be able to sort waste at the stations in many different locations; (ii) the manipulation controllers must generalize effectively to previously unseen objects, since people will deposit new and unexpected items into the bins; (iii) many of the objects that are encountered are especially difficult to manipulate, including deformable or complex shaped objects.
Examples are shown in Fig.~\ref{fig:diverse_waste_scenarios}.

We assign a reward to an episode when the robot picks up a misplaced object, moves it to the correct bin, and terminates the episode. After termination, we open the robot gripper and observe which bin the object landed in.
This reward structure results in an additional partial observability challenge, since the reward depends on where the object was grasped from (i.e., the robot can't simply ``cheat'' by picking up an already correctly sorted object from its bin and putting it back down), which motivates our use of recurrent architectures with memory as discussed in Sec~\ref{sec:method}.

\begin{figure*}[t]
    \centering
    \includegraphics[width=1.0\linewidth]{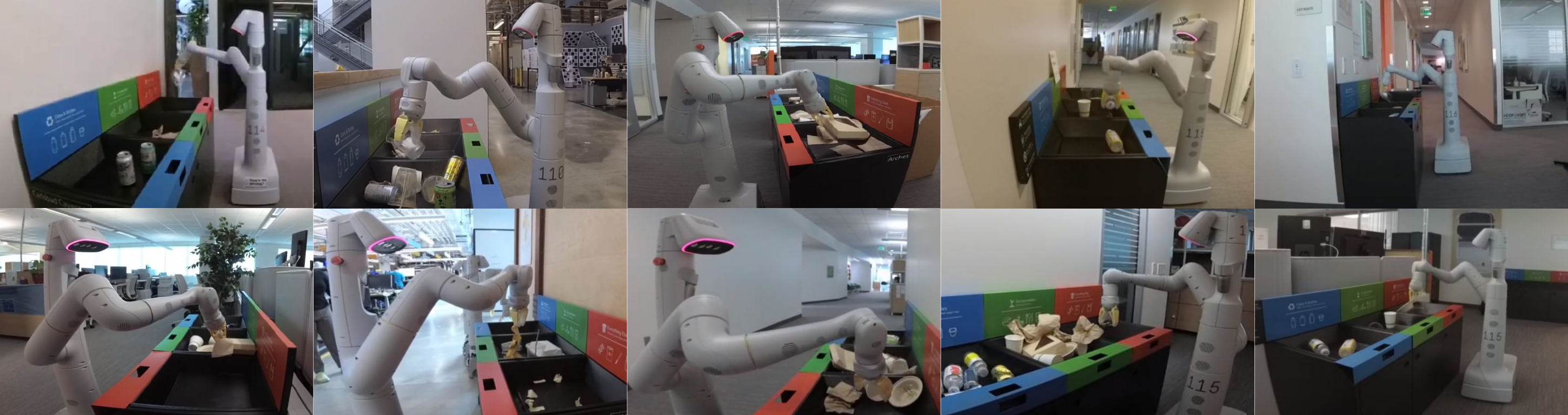}
    \caption{Example scenes from the deployment sites in 3 different buildings, showing the robot interacting with waste stations used by people over the course of their standard daily routines. Note the high variability in quantity and contamination levels.}
    \label{fig:site_deploy_examples}
    \vspace{-0.25cm}    
\end{figure*}

Iterations of our system, \algname, were deployed in three office buildings with 30 waste stations over the course of two years at various points in its development, resulting in 3803 autonomous station visits by the robots. Although the particular design of the system evolved over this time, this experience combined with more structured data collected in controlled settings, as well as simulated data, were included in the components of our final design, and the final evaluation scenarios used in our experiments were modeled on the challenges observed during deployment.
Fig.~\ref{fig:site_deploy_examples} shows a small subset of scenarios encountered by the robot during deployment.

\subsection{Robotic System Overview}
\label{sec:robot}

\begin{figure}[t]
    \centering
    \includegraphics[width=1.\linewidth]{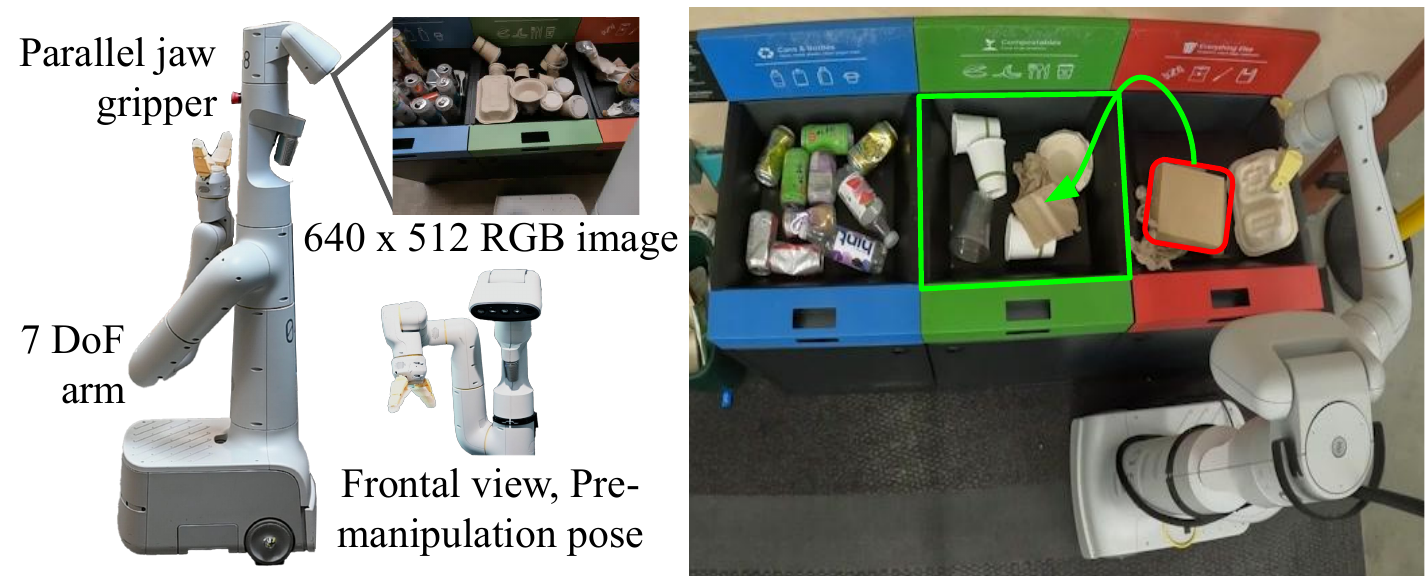}
\caption{The experimental platform. \textit{left}: Our mobile manipulator with a 7 degree-of-freedom (DoF) arm and a parallel jaw gripper.\\
\textit{right}: The sorting task demonstrated by an example: A compostable food container (red box) is misplaced in the landfill tray. Once the robot arrives at its initial state in front of the waste station with the arm above the station, it executes a trained or scripted policy that identifies misplaced objects and moves them to the correct bin. In the case of this example, the robot would receive a reward for moving  the food container into the compost tray (green box).}
    \label{fig:robot_and_task}
\end{figure}

We use a fleet of mobile manipulators with 7-degrees-of-freedom arms and parallel jaw grippers as shown in Fig.~\ref{fig:robot_and_task}.
For our problem, the state observations $\mathcal{S}$ for the robot corresponds to a $640 \times 512 \times 3$ RGB image observation from the robot's camera as well as a few proprioceptive signals that include the current tool height as well as the target tool pose and gripper aperture that the robot is moving towards. 
The action space $\mathcal{A}$ controls the whole body and consists of a target tool position and orientation and gripper aperture as well as target pose of the mobile base. 
Specifically, our action space consists of 10 dimensions: 3D position and orientation of the end-effector, gripper closedness, x and y position and yaw of the base and an added dimension for whether to terminate, each specified in the frame defined by the robot base and as deltas to the robot's current pose.
Our model chooses either an arm or a base motion at 1\si{Hz} or it chooses to terminate an episode. The commanded arm targets are mapped to straight-line Cartesian trajectories. Base targets are converted to a trajectory of two half-circles connecting the current and target base poses.

\section{\algfullname}
\label{sec:method}

Our method, \algname, incorporates end-to-end deep RL with data from simulation, data from the real world, and computer vision components to address the waste sorting task described in the previous section. While the basic deep RL methodology underlying our approach largely follows prior work~\citep{qtopt18,lee2022pi}, our full robotic learning system must address a number of important challenges that arise both with real-world application of deep RL in general, and with the waste sorting task in particular. First, deep RL requires overcoming exploration challenges that might necessitate impractical amounts of data collection when learning from scratch directly in the real world. We therefore employ several bootstrapping strategies to prime the learning process: we employ scripted policies that attain a low but non-zero success rate at object grasping (not necessarily the full sorting task) to collect initial object interaction data, together with simulated data collection. We also employ a multi-task curriculum learning strategy to bootstrap the complex sorting task with simpler tasks, such as grasping indiscriminately or grasping objects of specific types. This allows us to collect data both in simulation and in the real world, which we can incorporate into an offline RL method based on PI-QT-Opt~\citep{qtopt18,lee2022pi}. To address the partial observability of this manipulation task, we use recurrent models with memory (LSTMs) to represent the value function. Finally, since the waste sorting task requires broad generalization to new objects and semantic determination of the type of waste that each object represents (recyclable, compostable, or landfill), we additionally incorporate inputs from a computer vision system based on ShapeMask~\citep{kuo2019shapemask} to additionally boost the generalizability of our method, while retaining the benefits of end-to-end policy learning. We describe the bootstrapping procedure, data collection process, RL approach, and the vision system integrating in the subsequent sections.

\subsection{Bootstrapping}
\label{sec:bootstrapping}

In order to bootstrap our real-world \algname policy, we use three different and complementary sources of data: scripted policies, simulation, and multi-task RL objectives. We start this process by employing scripted policies in simulation and in the real world. The data collected from these is used to train RetinaGAN~\cite{ho2021retinagan}, which transforms simulated images to look more realistic. Equipped with this sim-to-real transfer tool we deploy policies learned in simulation in the real-world and run them side by side with scripted real-world policies. Since both of these policies encounter only occasional full-sort successes, we further employ multi-objective RL to enable faster bootstrapping. Below we describe these different components in more detail.

\textbf{Scripted policy.}
We start by implementing a simple scripted indiscriminate grasping policy (i.e., a policy that lifts any object) in simulation and in the real world. In each episode, the scripted policy would randomly detect an object using an off-the-shelf object detector (or in simulation ground-truth), plan the grasp pose, and generate the trajectories to reach the object. Even though the initial success rate of this policy is around $20\%$, it is enough to bootstrap simulation and start collecting data on the real robots. We start collecting data in a laboratory environment that we refer as \rc, where we deploy the scripted policy on $20$ robots that are located in front of sorting stations, where they continuously perform the task of indiscriminate grasping.

\textbf{Sim-to-real.}
Our simulation policy is initially bootstrapped using the same scripted policy as deployed in the real world. 
Once it gathers enough successes, we start reinforcement learning training in simulation. 
The simulation policy is trained with PI-QT-Opt~\cite{lee2022pi}.
Once a simulation policy is trained, to enable sim-to-real transfer, we transform simulated images to look more realistic using RetinaGAN~\cite{ho2021retinagan}, which is trained on real and simulation camera images. 
Even though this initial sim-to-real policy has a relatively low performance, it is good enough to simultaneously start deploying it in the \rc to provide an additional source of real-world data besides the scripted policy.

\textbf{Multi-task RL.}
Once we have good performance of indiscriminate grasping in both simulation and the classrooms, we start to encounter very occasional successes of the more complex tasks such as displacing an object (moving it from on tray to another) or even sorting (moving it to the correct target bin). Since the reward signals for the sorting task are still too rare, we introduce the final bootstrapping step that allows us to generate a large number of successful sorting examples. 
We take advantage of multi-task reinforcement learning where we introduce de-facto a curriculum of various task difficulties, the most difficult of which is the task of sorting waste. In particular, we train a sorting policy using a multi-task training strategy described in MT-Opt~\cite{mtopt21}. 
We therefore devise a total of 15 tasks, including ``indiscriminate grasping'', ``grasp recyclable'', ``grasp compostable'', ``grasp landfill'', ``indiscriminate from recycling'', ``indiscriminate from compost'', ``indiscriminate from landfill'', ``grasp misplaced recyclable'', ``grasp misplaced compostable'', ``grasp misplaced landfill'', ``displace object'', ``sort recyclable'', ``sort compostable'', ``sort landfill'' and finally ``sort''. Since the easier tasks experience more successes initially, they learn faster and lead to more successes of the more difficult tasks, which in turn bootstrap even more difficult tasks and so on. 
Once the multi-task policy has converged, we switch to only collecting data for the sort task and only training on the sort task for further improvements of the policy.

\subsection{Real-World Challenges and Data Collection}

Next, we go into detail on various design choices that were made to address the real-world challenges of collecting a large-scale dataset with autonomous robots sorting waste. 

\textbf{Scripted exploration.}
As described in the previous section, we start our data collection by running a scripted grasping policy that succeeds some of the time using off-the-shelf object detectors and grasp planning techniques. The goal of this policy is to generate enough data that, together with simulated data, will bootstrap the real-world RL policy. To accomplish this, the scripted policy needs to fulfill two requirements. First, it needs to encounter occasional successes that can be used to bootstrap the more powerful RL policy, which is described in Sec.~\ref{sec:bootstrapping}.
The second, more subtle requirement is that the distribution of actions for the scripted policy should be similar to that of the future RL policy, to provide sufficient coverage to support learning a more optimal policy with RL. 
To address this, we propose an action conversion mechanism and, together with the underlying script, we refer to this component as \textit{scripted exploration}.
To this end, the scripted policy we use for exploration first generates waypoints to reach for randomly selected objects proposed by an off-the-shelf object detector. It then uses these waypoints as ``attractors'' for a pseudo-value-function that is then fed to the same CEM process that we use to select actions with RL. That is, instead of optimizing the action via CEM to maximize its Q-value, we optimize it with CEM to minimize the distance to the waypoint, using the same proposal distribution that is used for running the RL policy. Since the stochastic CEM optimizer is imperfect, this leads to the same noise profile in both the exploration policy and the final RL policy. We present the details of the scripted exploration algorithm in the appendix.

\textbf{Autonomous data collection.}
Deploying robots for extended periods of autonomous operation and data collection, both in \rc and in the real world, is central to our data collection strategy. %
To prevent damage to the robot and its environment, while still allowing the robot to make contact with the world and manipulate it, we employ the following three strategies:
(1) When sampling potential actions for the arm and the base during the CEM phase of the algorithm, we restrict the samples for arm poses to a box spanning the waste trays and the area above and the orientation of the gripper to not be pointing upwards. CEM samples for movements of the base are constrained to a rectangular area in front of the waste station.\footnote{A visualization of the arm and base workspace constraints is shown in the supplemental material.} (2) We employ a controller that executes commands only in a \textit{best-effort} manner, meaning that trajectories are only executed to the point where any part of the robot (except the fingers) would collide with a voxel map representation of robot's direct environment, reconstructed from the robots depth sensors. (3) All motion is interrupted when a force threshold of 7\si{N} is exceeded at the wrist.

\begin{figure*}[t!]
    \centering
    \includegraphics[width=0.98\linewidth]{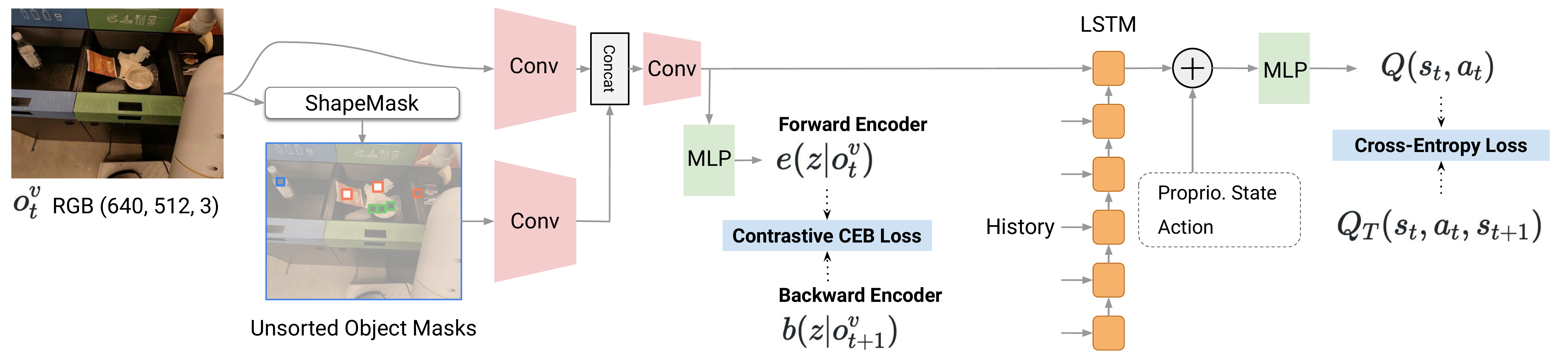}
    \caption{An overview of the network architecture. We encode RGB camera images and unsorted object masks convolutional layers. Our Q-function considers visual observations in the most recent 6 time steps, but the predictive information (CEB) auxiliary only considers the current image $o^{v}_{t}$ for the past $X$ and the next image $o^{v}_{t+1}$ for the future $Y$, in order to avoid information overlap between $X$ and $Y$.}
    \label{fig:architecture}
    \vspace{-0.25cm}
\end{figure*}

\subsection{Reinforcement Learning Flywheel}
\label{subsec:rl_flywheel}

Equipped with real and simulated data, we use deep RL to train an end-to-end policy that is directly optimized for reducing the contamination of the bins. 
Similarly to how we train our simulation policy, we leverage PI-QT-Opt~\cite{lee2022pi} to train the final policy on the complete dataset assembled from simulation and real world collection.
Deep RL allows us to not only distill the best possible policy out of the bootstrapping data, but also to enable the robot to improve continuously as it interacts with waste stations more and more. We refer to this iterative improvement as a \emph{data flywheel}: a continual process where the robot performs the waste sorting task, gathers more experience, and incorporates this experience back into the RL process to further improve its policy.

Since training and updating the policy after each sample across a large fleet of robots is impractical from a systems engineering perspective, we update the policy and deploy it to the fleet iteratively in batches.
For each iteration of the flywheel, we deploy an updated model to the robot fleet and collect a batch of data over a week of operation time. We train our model with PI-QT-Opt using all the data collected so far, including the newly added batch, and deploy the converged model back to the robot fleet and proceed with the next flywheel iteration.
Since we train our policy on real robots and we aim to deploy the best possible version of the policy at all times, we use an offline policy evaluation metric called OPC~\cite{irpan2019off} to find the most performant checkpoint of the current flywheel iteration.

\subsection{Neural Network Architecture}
\label{sec:architecture}

In Fig.~\ref{fig:architecture} we present the diagram of the neural network architecture of the Q-function that is learned with PI-QT-Opt. We feed two RGB images to two separate convolutional towers which are later concatenated and processed by another set of convolutional layers. The two images correspond to the current camera image as well as the object mask image that we describe in the next section.
After the last convolutional layers, we add a small multi-layer perceptron (MLP) to parameterize the forward encoder for predictive information learning.
We use the same convolutional architecture with moving average parameters of that for the current time step to encode the two RGB images of the next time step.
With the forward and backward encoders, we enforce the contrastive CEB loss described in Sec.~\ref{sec:prelims}.

To further aid with the long-horizon task of sorting waste, we enhance our policy by adding a set of LSTM layers to cope with the partial observability of the task. In particular, we employ a PI-QT-Opt-adjusted R2D2 approach~\cite{kapturowski2019recurrent} to pass the recurrent policy state from data collection to the replay buffer and further to the trainer.
Towards the end of the neural network processing, we add proprioceptive information as well as the action that is being scored by the Q-function. These, together with the previous signals are processed by an extra set of MLP layers to output the final Q-value.

\subsection{Boosting Generalization with Pretrained Object Masks}

While the simulated and real-world datasets can provide our waste sorting policy a sufficient level of performance to continually collect useful data, diverse real-world sorting scenarios necessitate a very high degree of generalization due to the variety of objects that the robot is likely to encounter. Furthermore, the robot must be able to correctly determine the type of each object (compostable, recyclable, or landfill). Therefore, to further boost generalization, we integrate a pre-trained computer vision model that is further fine-tuned on labeled data for the sorting task.

We use a pre-trained one-stage segmentation model to predict panoptic segmentation masks (a combination of instance and semantic masks). The model consists of an EfficientNet-B1~\cite{Tan2019EfficientNetRM} backbone and separate convolutional heads for object-wise classification, box regression, and mask prediction. We also leverage mask priors to help generalize to different object shapes as in ShapeMask~\cite{kuo2019shapemask}.
The backbone is first pretrained on ImageNet~\cite{imagenet} through the noisy-student setup~\cite{Xie2019SelfTrainingWN}, and then the entire model, including the mask priors, is finetuned on a separate perception dataset consisting of 30k examples labeled with panoptic masks, all collected from the robot during operations in the same waste sorting task setup.
The list of object classes used for semantic segmentation is a superset of the waste objects seen during sorting.

Once the object mask predictor is trained, we incorporate the information it provides into the policy network. To achieve this, we create an extra image with a dot at the center of every object that is currently misplaced. The color of the dot indicates which bin the object should be sorted into. This image is fed to the network as an extra input channel concatenated to the current RGB image as shown in Fig.~\ref{fig:architecture}.

This design presents a number of advantages. We incorporate knowledge from pretraining on computer vision datasets,
enabling the robots in principle to classify and sort objects that they have not interacted with.
Further, it allows the RL policy to focus on the manipulation aspects of the sorting task (e.g., how to singulate and grasp an object) rather than the semantic classification, which can be achieved more efficiently using supervised learning. 
At the same time, in contrast to more standard approaches that decompose localization, planning and control into separate modules, our deep RL approach is still end-to-end, in the sense that the Q-function still operates on raw images and actions and can use them in whatever way it needs to maximize task performance, while additionally receiving the object mask as a supplementary input.

\section{Real-World Experiments}
\label{sec:experiments}
In our evaluation, we seek to answer the following questions: (1) How does the performance of \algname change with various amounts of data in terms of in-distribution performance as well as generalization to new sorting scenarios? (2) What is the impact of various design decision on the final performance of \algname? (3) What are the overall trends and performance metrics from deploying \algname in real office buildings?

\subsection{Experimental Setup}
\label{sec:exp_setup}

\begin{figure*}[t]
    \centering
    \includegraphics[width=1.0\linewidth]{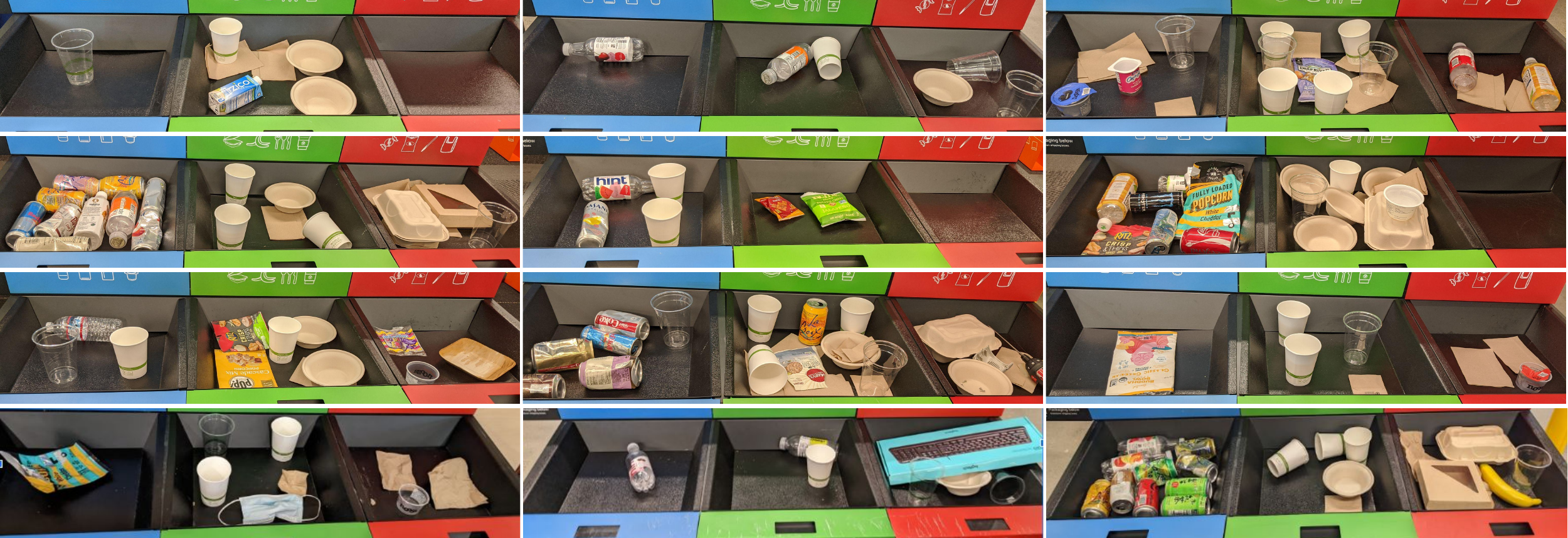}
    \caption{Waste scenarios used for evaluations. Top 3 rows show the 9 in-distribution scenarios. The bottom row shows the held-out-scenes, containing objects previously seen neither in the real world nor in simulation, such as the keyboard, banana and face-mask.}
    \label{fig:waste_scenarios}
    \vspace{-0.4cm}
\end{figure*}

\begin{figure}[ht]
    \centering
    \includegraphics[width=1.0\linewidth]{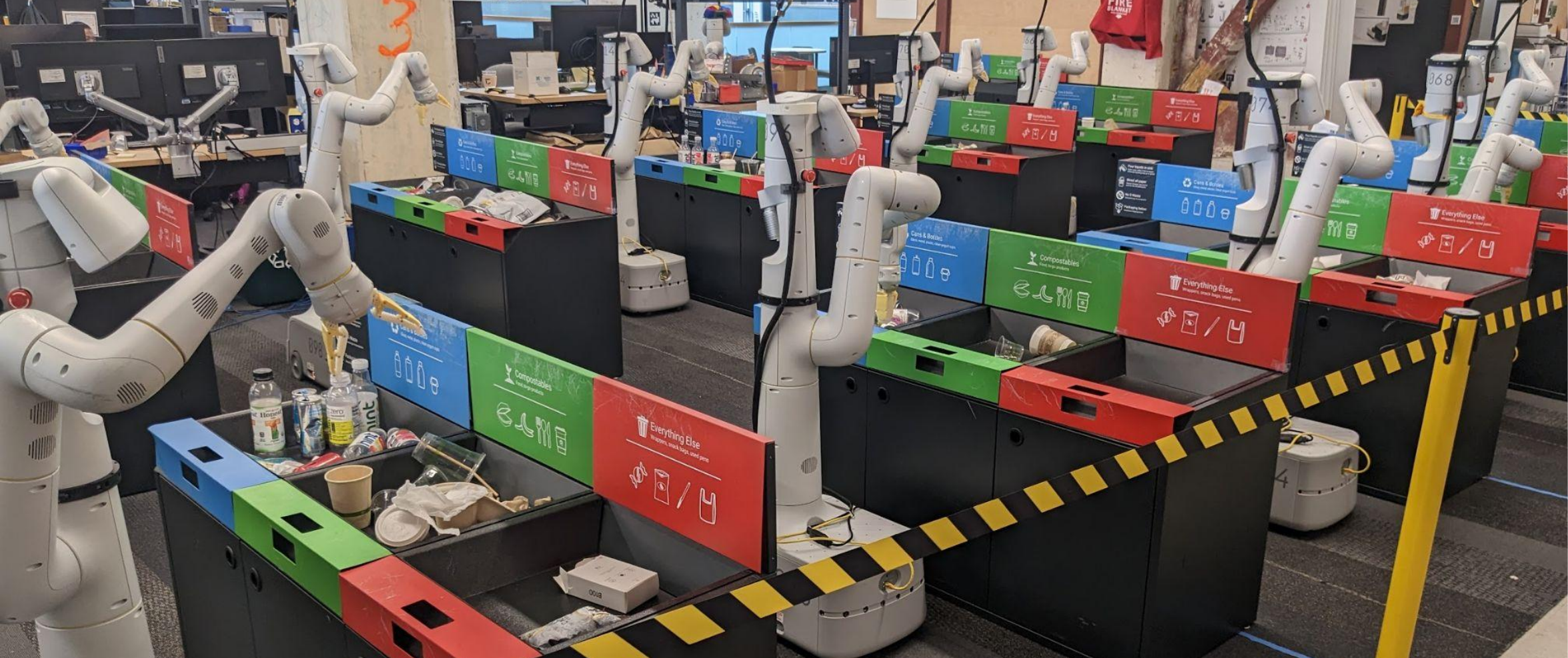}
    \caption{The robot classroom, a controlled setting for repeatable evaluations. 20 robots continuously collect data at 20 waste stations.}
    \label{fig:classroom}
    \vspace{-0.3cm} 
\end{figure}

For data collection in the \rc, we use a fleet of 20 robots that continuously sort waste from 20 waste stations (see Fig.~\ref{fig:classroom}). We randomly place objects into the waste stations and shuffle them several times throughout the day to increase contamination. After an initial bootstrapping phase (see Sec.~\ref{sec:bootstrapping}) we deployed our flywheel and collected $540$k episodes in the \rc over the course of 4 months. Over the course of the project, we also deployed various iterations of our method to 23 robots servicing 30 waste stations across three operational office buildings, which we refer to as the \textit{deployment site}, which we describe in detail in the appendix. Waste stations at deployment sites are filled with waste entirely by people that ordinarily work in these buildings. We gathered $32.5$k episodes from our deployment and added them to the overall training set. Note that data throughput was significantly higher in the \rc, where robots could sort waste continually, whereas the available unsorted waste at the \textit{deployment sites} varied drastically over time (particularly during stages of the project that overlapped with the COVID epidemic).
For repeatable experimentation, we identified 9 challenging waste scenarios from the deployment site and additional 3 held-out scenes which contain objects that the robot has not interacted with in the training data (see Fig.~\ref{fig:waste_scenarios}).
A waste scenario prescribes how many instances of each object class are placed in which of the three bins before the robot starts sorting. 
The placement of objects within each bin are randomized. 
We perform extensive evaluations on the two sets of waste scenarios.  Each robot has a maximum of 20 attempts to sort all the objects in each scenario, and the scenarios have between 2 and 9 initially misplaced objects. We report the sorting success rate over 2 rounds for a total of 360 attempts (20 attempts $\times$ 9 scenarios $\times$ 2 rounds).

\subsection{Quantitative Evaluation of \algname}
\begin{figure}[t]
    \centering
    \includegraphics[width=0.9\linewidth]{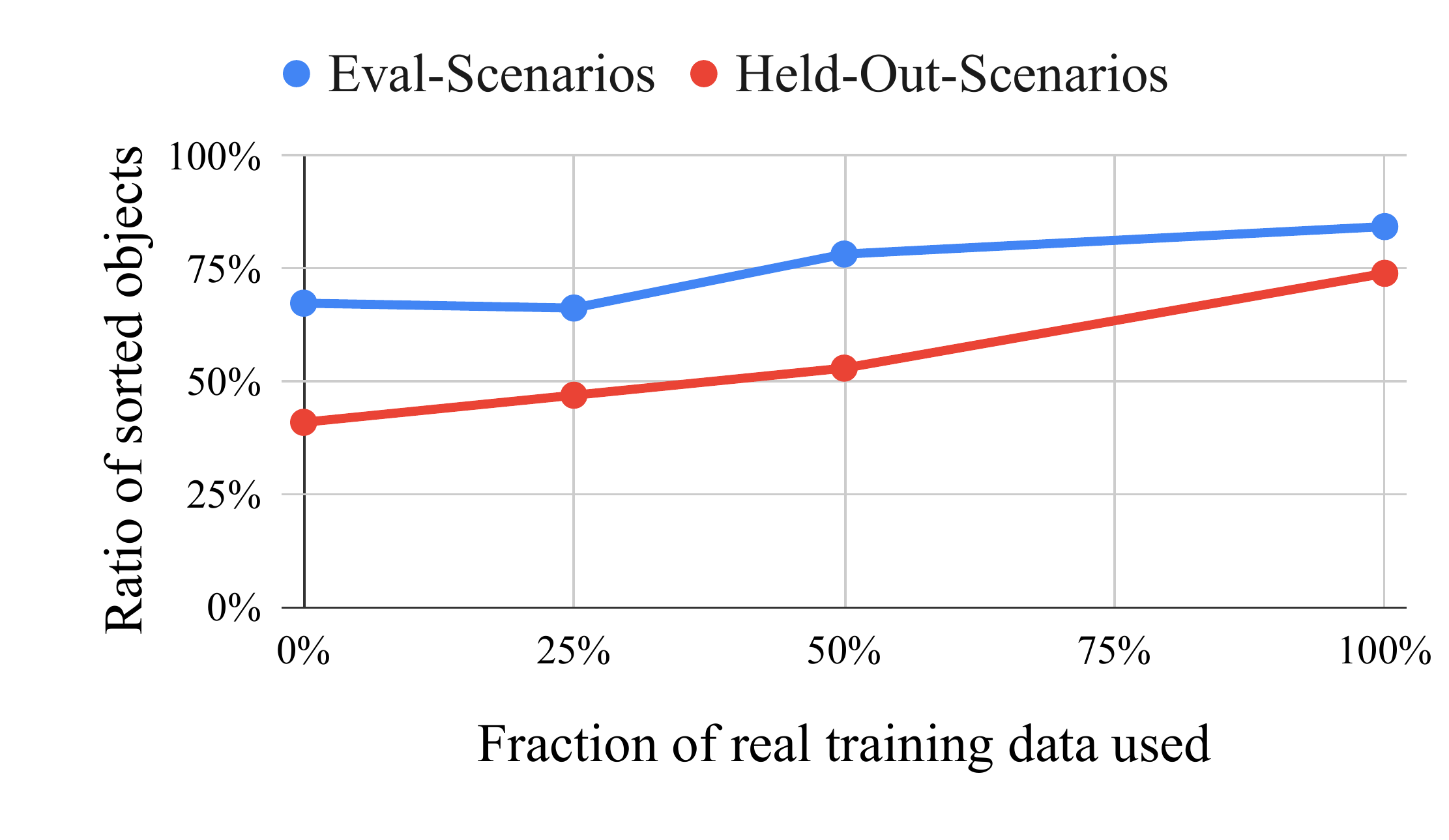}
    \caption{Waste sorting performance when using 0\%, 25\%, 50\%, and 100\% real world training data on the evaluation scenarios (blue) and the held-out scenarios (red). 0\% represents training \algname only in simulation. As expected, the performance scales with the amount of training data used, is slightly better on the evaluation scenarios, but also generalizes and improves to excellent performance on the held out scenes.} %
    \label{fig:data_ablation}
    \vspace{-0.25cm}    
\end{figure}
Since our system \algname is deployed at scale over the course of multiple months, including deployment in actual office buildings with constantly changing trash scenarios, it is challenging to perform a clean set of experiments delineating the influence of different components of the system. We therefore use the \rc setup for ablations and experiments, using the scenarios described in Sec.~\ref{sec:exp_setup} and shown in Fig.~\ref{fig:waste_scenarios}.

To answer our first question: \textit{How does the performance of \algname change with various amounts of real world data in terms of in-distribution performance as well as generalization to new sorting scenarios?} we perform a set of evaluations where we vary the amount of real-world data included in training the model.
In particular, we perform training of \algname in iterations in the \rc, where each iteration involves using the policy from the previous iteration to collect data for several days. 
A new policy is then trained on all the data, including the latest iteration, and if performance improves over the previous policy then it is used as the collection policy for the next iteration. 
For data collection, we set up scenarios with the same objects as the in-distribution evaluation scenario, but with object numbers and configurations randomized, and we additionally report performance on unseen scenarios. Robots are allowed to autonomously operate with occasional reshuffling of the objects by human operators. All episodes are labeled by humans (via an image labeling interface similar to ones used for other image labeling work) to mark if an object is sorted or not, and these binary labels are directly used as the RL reward.

We present the results in Fig.~\ref{fig:data_ablation}. In-distribution performance is averaged over nine evaluation scenarios, and generalization performance is averaged over three held-out scenes, as described in Sec.~\ref{sec:exp_setup}. The policy steadily improves with more real-world data, with 84\% of the objects correctly sorted for the final policy. We observe the biggest improvement in the held-out scenarios when switching from $50\%$ of the data to $100\%$, indicating that more data has a substantial impact on the generalization of \algname.

\subsection{Qualitative Evaluation of \algname}

\begin{figure*}[ht!]
    \centering
    \includegraphics[width=1.0\linewidth]{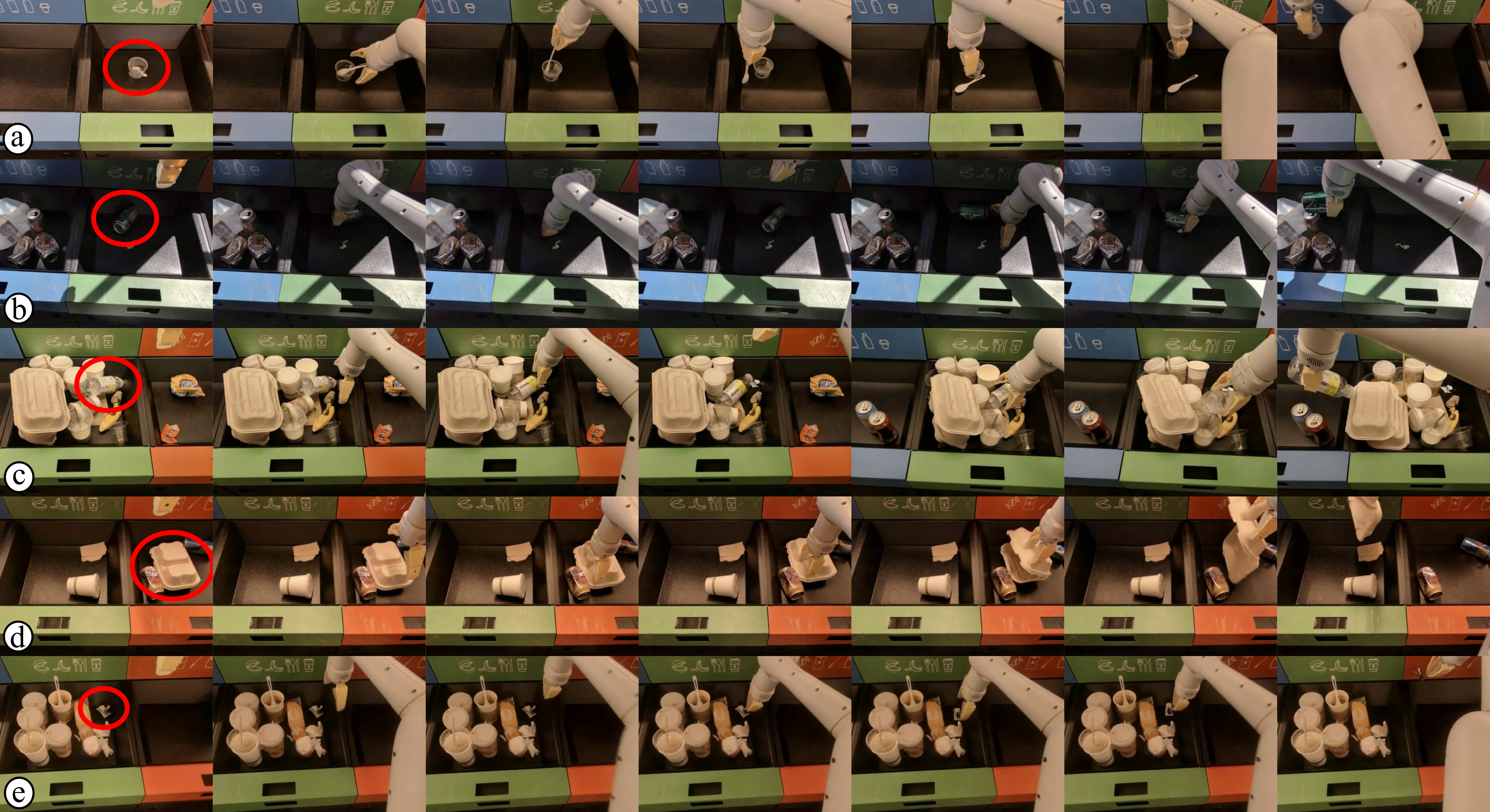}
    \caption{Five successful sorts illustrating the dexterity of the learned policy: (a) Removing the compostable spoon (red circle) and correctly sorting the yogurt cup. (b) Sorting a soda can (red circle) under poor lighting. Note how at first the can is occluded and the robot lifts its gripper to make it visible again. (c) Sorting a plastic bottle (red circle) after a few re-grasps. (d) Sorting a large lunch box. (e) Pinching and sorting a small candy wrapper (red circle).}
    \label{fig:sort_examples}
\end{figure*}

In order to illustrate the difficulty of some of the encountered sorting scenarios, we present a few examples in Fig.~\ref{fig:sort_examples}. These include a number of contact-rich, difficult-to-accomplish manipulation behaviors. These are difficult not only in terms of manipulation, but also the semantics of the task (e.g., Fig.~\ref{fig:sort_examples}a), where the policy needs to distinguish that the object inside a container is correctly sorted already, while the container itself needs to be moved to a different bin.
In Fig.~\ref{fig:sort_examples}b, we show an example with poor lighting, which do not prevent our policy from correctly sorting the object. We also demonstrate an emerging re-grasping behavior of our policy (e.g., Fig.~\ref{fig:sort_examples}c), as well as the ability to cope with very large (Fig.~\ref{fig:sort_examples}d) and very small objects (Fig.~\ref{fig:sort_examples}e).

\subsection{Ablations of \algname and Design Decisions}
\begin{figure}[t]
    \centering
    \includegraphics[width=1.0\linewidth]{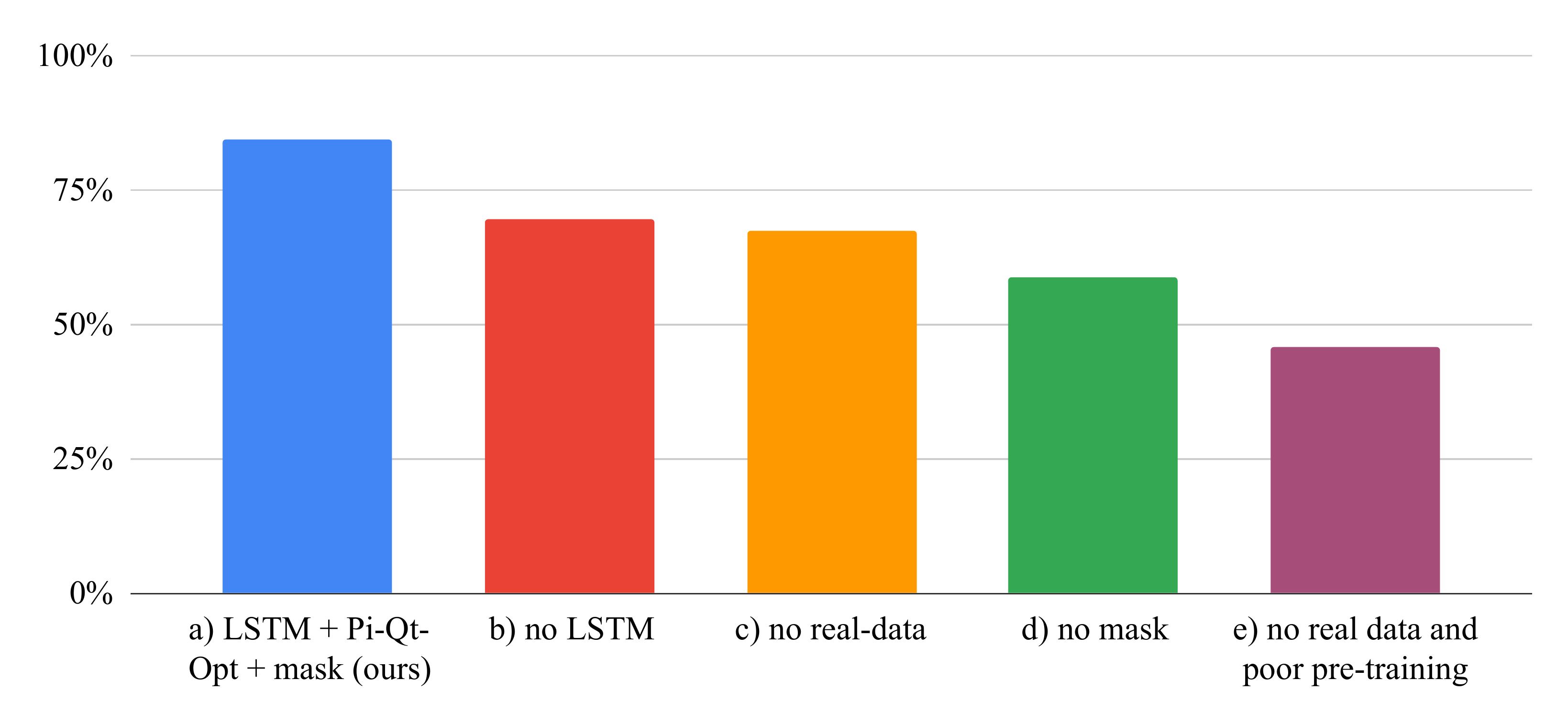}
    \caption{Ratio of sorted objects of ablations of our approach. a) our method, RLS b) our method without LSTM c) trained with sim data only c) without mask-conditioning d) trained with sim data only and a poorly performing bootstrap model.}
    \label{fig:feature_ablation}
    \vspace{-0.4cm}
\end{figure}
Next, we study the second question: \textit{(2) What is the impact of various design decision on the final performance of \algname?}.
We evaluate the performance of variations of our method on the 9 waste scenarios and report ratios of objects sorted in Fig.~\ref{fig:feature_ablation}. 
We observe the biggest negative impact, $45.66\%$ objects sorted, when we train our policy purely on simulated episodes without including any real data, and when in addition we choose a poor initial policy to bootstrap the training process.
If we use the original bootstrap model and no real data, we see a performance of $67.39\%$.
We observe a performance of $58.7\%$ when we remove the object mask, forcing the policy to rely on only the RGB input, indicating the value of introducing auxiliary semantic knowledge from the masks.
If instead we remove the LSTM architecture and fall-back to a memory-less architecture we see a performance of $69.57\%$. Note that for this baseline, we filter out episodes where the robot incorrectly grasped an object that was not misplaced, so the drop in performance is not due only to the inability to determine which objects were already sorted.
Finally, our method, \algname, with all design decisions included, sorts $84.35\%$ of the objects.
Overall, we see that the design choices made for \algname are important to achieve good performance on a challenging real world task as trash sorting.

\subsection{Evaluating the Real-World Deployment of \algname}
\label{sec:deployment_eval}

We deployed our system, \algname, in three office buildings using a fleet of 23 robots servicing 30 waste stations over the course of 14 months.
Based on this deployment, we gathered $32.5$k waste sorting episodes and visited the sorting stations a total of $3803$ times. The unstructured nature of this task makes it difficult to perform rigorous head-to-head comparisons, but we report some statistics of this deployment below.

First, to provide an intermediate evaluation between the more structured \rc and the fully unstructured deployment, we picked one waste station in each building and set up a scene analogous to our \rc evaluation scenario in each one. In this case, the results show a negligible difference, with a $92.7\%$ success rate in \rc and $93\%$ in the office buildings, confirming that our policy generalizes to different locations.

Next, we fully deploy \algname ``in the wild,'' where our robot fleet needs to sort previously unseen waste in novel waste sorting scenarios that are created by the occupants of these buildings. We deploy our system over three time periods in 2021 and 2022, each one of them spanning between 100 and 170 days. Because of the global pandemic, the first two deployments experience much lower amounts of waste in the sorting stations, and therefore fewer waste station visits. We summarize our results in Table~\ref{table:in_the_wild}.

\begin{figure}[t]
    \centering
    \includegraphics[width=1.0\linewidth]{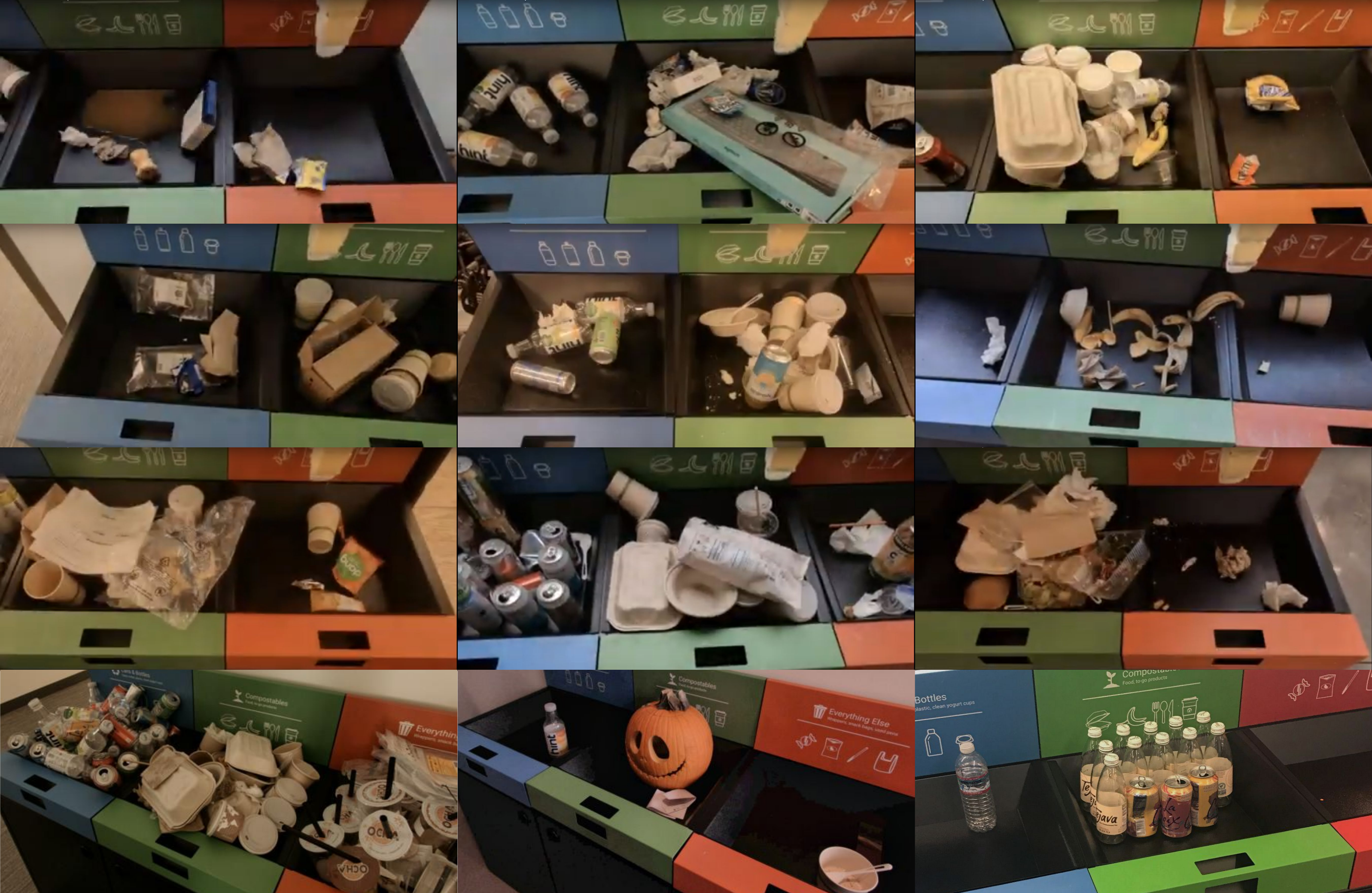}
    \caption{Example waste scenarios observed during site deployment. We observe a significantly wider range of objects such as various packaging materials, large plastic wrappers, very small torn up papers/candy wrappers, glass bottles, spilled liquids, and perished food. Also, we observe much more diverse object configurations, trays are densely packed and objects are carefully piled and possibly overhanging.}
    \label{fig:diverse_waste_scenarios}
    \vspace{-0.4cm}
\end{figure}

The contamination reduction in Table~\ref{table:in_the_wild} is calculated as a ratio between the \emph{weight} of all the initially misplaced objects in the bins and the weight of the misplaced objects at the end of \algname runs.
We measure the contamination reduction to initially be $53\%$ during the first deployments, which experienced lower traffic because of the pandemic. The contamination reduction drops to $39\%$ once there are more people in the office buildings, when our system experiences much more diverse sorting scenarios.
To visualize the diversity of the encountered scenarios, we show some examples in Fig.~\ref{fig:diverse_waste_scenarios}. These include overflowing bins (last row, left column), objects that are too large for our robot to possibly grasp, such as keyboard packaging (first row, middle column), and tightly packed objects that are difficult to manipulate (last row, last column).
These examples illustrate the gap between the real-world ``in the wild'' distribution of sorting scenarios and the distribution experienced in the \rc, where \algname achieves a high sorting performance. However, in all cases, the \algname policy led to significant reductions in contamination, resulting in a considerable reduction in the amount of waste that would otherwise have been improperly sent to a landfill or contaminated a recycling batch.

\begin{table}
\centering
  \begin{tabular}{@{}lcc@{}}
\hline
  Deployment & Waste station & Contamination\\
days (dates) & visits & reduction \\
  \hline
  165 (11/2021-4/2022) & 15 & 53\%\\
  160 (8/2021-2/2022) & 51 & 52\%\\
  104 (5/2022-8/2022) & 277 & 39\%\\
\hline
  \end{tabular}
\caption{Statistics from the three deployments in three office buildings that span $\sim2$ years. The first two rows showcase fewer waste station visits because of the global pandemic, which resulted in lower amounts of trash in the waste sorting stations in the office buildings.}
\label{table:in_the_wild}
\end{table}

\section{Limitations and Conclusions}
\label{sec:conclusions}

In this paper, we presented a robotic manipulation system based on reinforcement learning that sorts waste at waste stations in office buildings, described a large-scale deployment of this system, and presented a quantitative evaluation based on the experience from this deployment. Our approach aims to illustrate how deep RL methods can be integrated into a robotic manipulation system that is feasible to deploy in the real world and able to benefit from real-world data, simulation, and improved generalization from computer vision components. The policy is bootstrapped from a combination of simulated data and experience collected with a scripted policy, from where it is further improved through additional data collection in \emph{robot classrooms} and deployment. The training process employs a curriculum that integrates multi-task learning with simpler tasks to bootstrap the complete sorting task, uses a memory-based LSTM architecture, and integrates additional inputs from an object mask trained on labeled data to boost generalization. We ablate the individual components of our system and evaluate how its performance improves with more data, including on held-out test scenarios.

One limitation of our system is that it still relies heavily on experience gathered in the controlled \emph{robot classroom} settings, rather than entirely using data from real-world deployments. We found the deployments to be highly variable, and also harder for obtaining a steady volume of useful experience, as real-world waste was deposited into the waste stations at different rates during the day and week, resulting in lower data collection throughput than we could obtain in the \emph{classrooms}. 
We hope that the demonstration of large-scale real-world robotic RL presented in this paper can serve as a prototype for future works that will address this and other future challenges and deploy complete robotic manipulation systems that incorporate RL to enable real-world improvement and generalization.

\subsection*{Acknowledgements}
The authors would like to thank Mohi Khansari, Cameron Tuckerman, Stanley Soo, Justin Vincent, Mario Prats, Thomas Buschmann, Joséphine Simon, Jarrett Lee, Kalpesh Kuber, Meghha Dhoke, Christian Bodner, Russell Wong and the entire Everyday Robots team for their help and support in various aspects of the project.

\bibliographystyle{abbrvnat} %
\bibliography{references}

\clearpage
\newpage

\appendix
\label{sec:app}

\subsection{Scripted exploration}
We present the scripted exploration algorithm in Alg.~\ref{alg:scripted_expl}. Note that the action length and noise of $s_min$ depend on the initial parameters of CrossEntropyMethod (line~\ref{alg:scripted_expl:cem}). In particular, we run only 2 iterations of the method for faster computation. We use the same CrossEntropyMethod parameters with our learned critic (Fig.~\ref{fig:architecture}) for inference and training. This ensures that generated action samples from the script (line~\ref{alg:scripted_expl:script}) and our learned critic resemble the same action distribution. 

\begin{algorithm}[!ht]
\DontPrintSemicolon
 \SetKwFunction{genposes}{GenerateTargetWaypoints}
 \SetKwFunction{dist}{EuclideanDistance}
 \SetKwFunction{argmin}{argmin}
 \SetKwFunction{random}{Random}
 \SetKwFunction{robotmove}{Robot.Move}
 \SetKwFunction{CEMSampler}{CEMSampler}
 \SetKwFunction{cem}{CrossEntropyMethod}

 \SetKwProg{Fn}{Def}{:}{}
  \Fn{\genposes{}}{  \label{alg:scripted_expl:script}
        \If{IsSimulation} {
           object\_poses $\gets$ simulation ground truth
        } \Else {
           object\_poses $\gets$ object detector
        }
        $w_{approach} \gets$ \random(object\_poses)
        
        $w_{grasp} \gets w_{approach}$ with gripper closed
        
        $w_{lift} \gets w_{grasp}$ with lift $20cm$ along z-axis
        
        \KwRet $w_{approach}, w_{grasp}, w_{lift}$
  }
  \Fn{\cem{critic}}{  \label{alg:scripted_expl:cem}
    Initialize Gaussian, $\mu, \sigma$

    Initialize number of iterations, $I = 2$

    Initialize batch size, $N = 128$

    Initialize number of elites, $M = 13$

    \For{\_ in I}{
        $\hat{s}_{1:N} \sim \mathcal{N}(\mu,\,\sigma^{2})$

        $\hat{d}_{1:N} \gets critic(s_1), \dots, critic(s_N)$
        
        $d_1, s_1, \dots, d_N, s_N \gets rank(\hat{d}_1, \hat{s}_1, \dots, \hat{d}_N, \hat{s}_N)$

        $\mu = \frac{1}{M} \sum_{s_{1:M}}{s_i};~ \sigma^2 = \frac{1}{M} \sum_{s_{1:M}}{(s_i - \mu)^2}$
    }
    
    \KwRet $s_1, d_1$
    
  }

  Initialize waypoint threshold, $d_{threshold}$

    \For{w in \genposes$()$} {
        $d_{min} \gets \inf$

        \While{$d_{min} \geq d_{threshold}$} {

            $s, d \gets CEM(\dist(w, \cdot))$

            \If{$d \leq d_{min}$} {
                
                $d_{min} \gets d$ \;
                    
                $s_{min} \gets s$ \;

            }
            }
            \robotmove($s_{min}$) \;
        }
 
\caption{Scripted Exploration Policy}
\label{alg:scripted_expl}
\end{algorithm}

\subsection{Predictive information preliminaries}

\emph{Predictive Information}~\cite{bialek1999predictive}, the mutual information between the past and the future, has been an effective auxiliary for large-scale real-world RL~\cite{lee2022pi}.
From here on, we will denote the past by $X$ and the future by $Y$.
In an MDP, the past $X$ refers to what has been observed by the agent; the future $Y$ refers what will happen in the future process.
\citet{lee2020predictive} argues that a learned representation $Z$ of the predictive information should be compressed with respect to $X$, based on the observation of \citet{bialek1999predictive} that $H(X)$ grows faster than $I(X;Y)$.
We follow \citet{lee2020predictive} to learn the representation $Z$ with the Conditional Entropy Bottleneck (CEB)~\cite{fischer2020conditional}, using the same contrastive variational bound:

\begin{align*}
CEB &\equiv \min_Z \beta I(X;Z|Y) - I(Y;Z) \\
&\leq \min_Z \mathrm{E}_{x,y,z \sim p(x,y)e(z|x)} \beta \log \frac{e(z|x)}{b(z|y)} \\
&- \log \frac{b(z|y)}{\frac 1 K \sum_{k=1}^K b(z|y_k)} \label{eq:ceb}
\end{align*}
where $(x,y)$ are sampled from the data distribution, $p(x,y)$, $e(z|x)$ is the learned \emph{forward encoder} distribution, $b(z|y)$ is the learned variational \emph{backward encoder} distribution, $\beta$ is a Lagrange multiplier that controls how strongly compressed the learned representation $Z$ is, with smaller $\beta$ corresponding to less compression, and $K$ is the number of examples in a mini-batch during training.
We choose $e(z|x)$ and $b(z|y)$ to be parameterized von Mises-Fisher distributions as in \cite{lee2021compressive}.

\subsection{Fleet Deployment}
\label{sec:deployment}

Once our policy achieves a satisfying performance in the \rc, we deploy it in three real office buildings using a fleet of 23 robots.
The robots use SLAM to localize themselves in the new building and navigate to a waste sorting station in their proximity. Once the robot is in front of the bin, we switch on our end-to-end \algname policy to perform the waste sorting task.

Since many of the waste sorting scenarios that we encounter in office buildings have not been seen previously and are likely to be seen only once due to the non-stationary nature of the task, and the overall throughput of data collection is limited (since people only deposit a limited amount of waste in the bins), we recreated the most commonly seen real-world scenarios in the \rc to enable the robots to practice a realistic scenario for more trials.
We periodically update these scenarios in the \rc to reflect the most common waste sorting challenges encountered during the deployment.
We use the data gathered during the deployment phase to train and improve all aspects of the \algname system: the classifier used for the object masks, RetinaGAN used for sim-to-real transfer as well as the PI-QT-Opt policy itself. To ensure that the most performant policy is deployed in office buildings at all times, we run the data flywheel described in Sec.~\ref{subsec:rl_flywheel}, which includes one week worth of data. While the bulk of our evaluations were conducted in the \rc using scenarios recreated from real-world deployment (in order to allow systematic comparisons and repeatability), we discuss some performance measurements that we recorded during deployment in Section~\ref{sec:deployment_eval}.

\subsection{Neural network parameters and training}

The critic network presented in Figure~\ref{fig:architecture} has a total of 1.43M learnable parameters. The forward and backward encoder for the predictive information loss each have 67k parameters. The architecture and parametrization largely follows~\cite{lee2022pi}. In addition, the added LSTM is implemented by a ConvLSTM2D layer with 64 $3\times3$ filters.

During training, we use Momentum optimizer with a learning rate of 0.0095 and momentum of 0.924; the batch size is 4096, training on a $4\times4$ slice of a TPUv3 pod (batch size 256 per chip).
As described in Section~\ref{subsec:rl_flywheel} we chose a checkpoint from training by measuring OPC~\cite{irpan2019off} on a held out dataset. To that end we train the model until OPC stabilizes, which typically happens well within the first 1 million training steps, and pick the checkpoint with the best OPC value (typically somewhere between 300k and 500k steps).

Overall, the training infrastructure for \algname consists of a large scale distributed system: 5000 simulator jobs continuously pick up new policy checkpoints and use it to gather new experience. 2000 log replay jobs read episodes collected on the real robots from disk. Both these data sources push episodes to a distributed set of 20 replay buffers.
A separate set of jobs pulls from those buffers mixing real and simulated data at a ratio of 1:1, computes Q-value targets with the Bellman equation and trains new model checkpoints. In parallel, sets of simulators evaluate model performance in simulation, and another set of jobs calculates the OPC metric for each new checkpoint.

The stability of the data flow through this system is crucial for the performance of the resulting trained policy, but not trivial to achieve in a distributed system where every part can be randomly preempted due to resource constraints or machine failures. We thus spent considerable time fine-tuning it. 
We find that the most crucial part is to ensure that the training does not overfit to the experience in the replay buffers at any point in time, when data production slows down due to partial outages of the feeding jobs. The replay buffers are configured to hold a maximum of 2500 episodes, removing old episodes in FIFO manner when the buffers are full and additional data gets pushed.  
Sampling from the buffers occurs by drawing batches randomly across all samples currently in the buffer. To prevent overfitting we set a maximum re-use for each sample of 40 before we drop it from the buffer. Additionally we only allow sampling when the buffer contains at least 500 samples; otherwise the whole system waits until new data arrives.

\subsection{Workspace safety constraints}

\begin{figure}[ht!]
    \centering
    \includegraphics[width=1.0\linewidth]{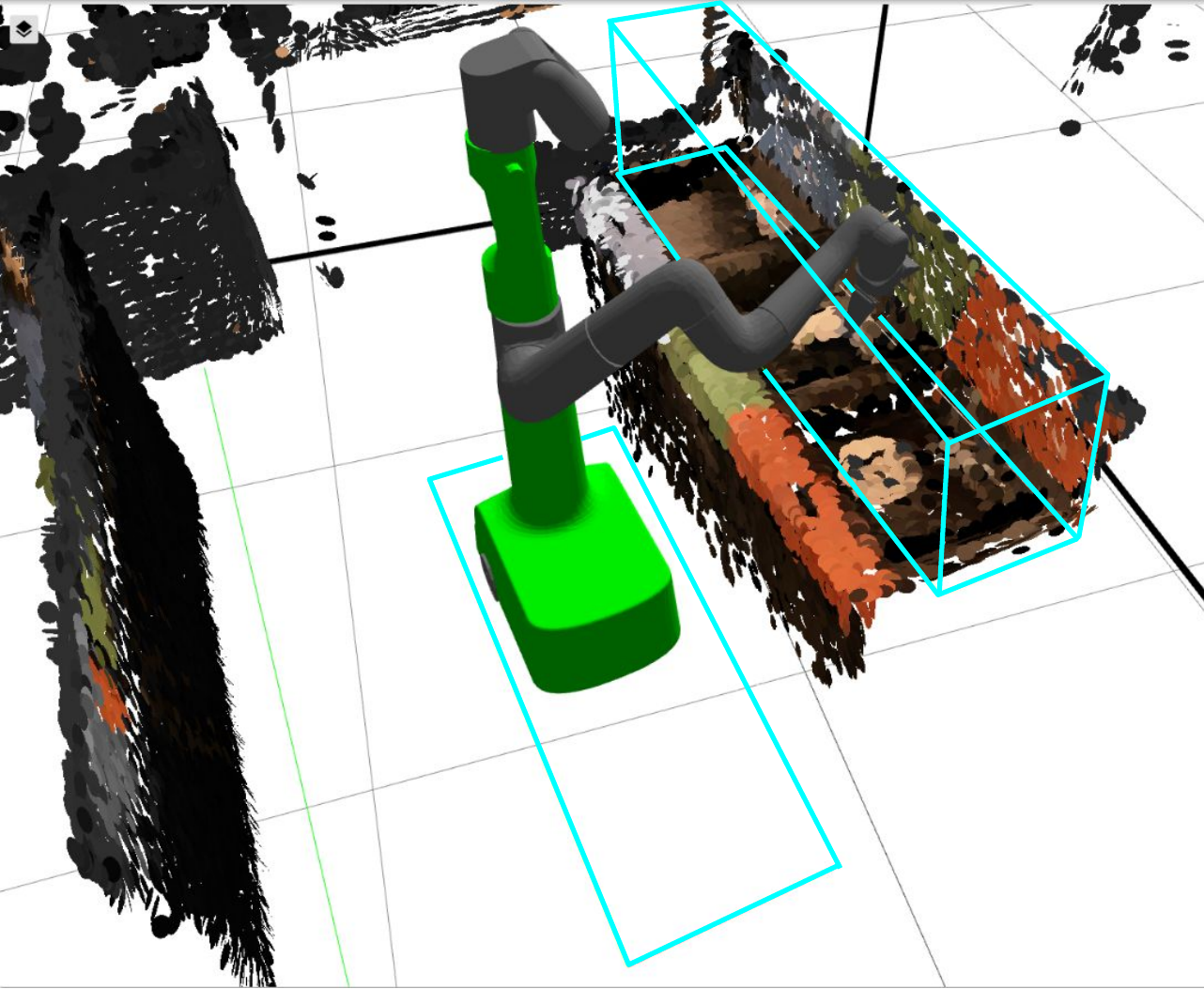}
    \caption{A visualization of the robot and action-space boundaries. We span a box-constraint for the end-effector above the waste stations that allows the robot to explore sorting strategies while excluding non-productive poses. Similarly, for the base we define an area in front of the waste station that allows the robot to reach all parts of the waste station.}
    \label{fig:action_bounds}
\end{figure}

Figure~\ref{fig:action_bounds} shows a visualization of the workspace safety constraints for the robot's base and end-effector. Random samples around the robot's current pose are clipped to these constraints before being scored by the policy Q-value function in the CEM process.

\subsection{Simulation details}

\subsubsection{RetinaGAN for visual domain transfer}

\begin{figure}[tb]
    \centering
    \includegraphics[width=1.0\linewidth]{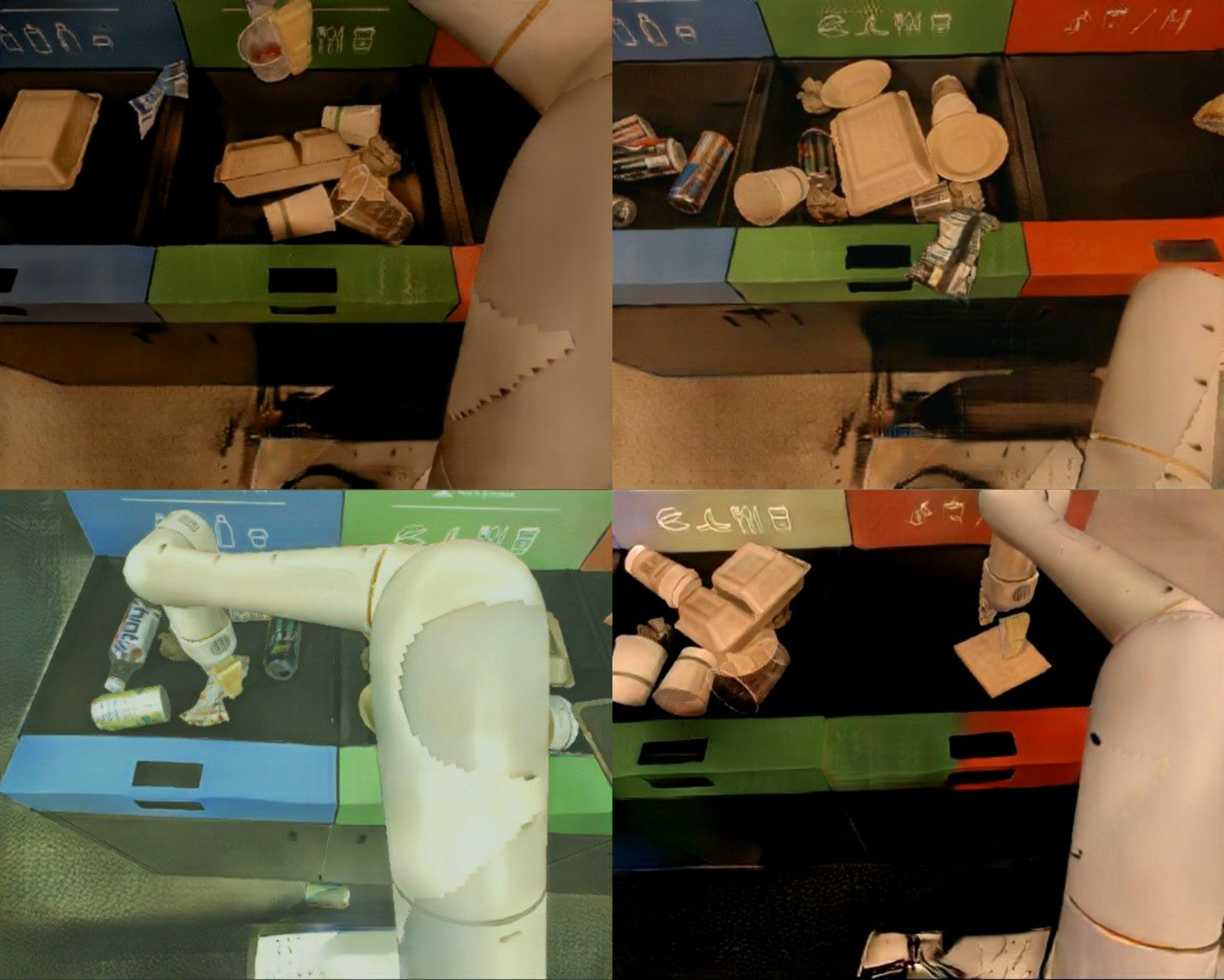}
    \caption{Example 640x512 images generated using our simulator and adapted with  RetinaGAN~\cite{ho2021retinagan}.}
    \label{fig:retinagan_examples}
\end{figure}

As described in Section~\ref{sec:bootstrapping}, following previous work we employ RetinaGAN~\cite{ho2021retinagan} to narrow the visual simulation-to-real domain gap, by applying an adapter GAN model to the images from our simulation. This approach greatly reduces the burden of creating high fidelity models and textures and the computational cost of complex rendering, allowing us to run our large scale simulations on machines without GPUs or other accelerators in the cloud. Figure~\ref{fig:retinagan_examples} shows examples of images from our simulation adapted with RetinaGAN.

\subsubsection{Deformable objects}

For many of the objects we deal with, like cans and bottles, it is sufficient to model and simulate them as fully rigid, without introducing a large domain gap in the physical behavior. For others though, like for instance empty chips bags or other snack wrappers, we found it crucial to model and simulate them as deformable. If chips bags were modeled as rigid we would have to decide on a rigid shape. If we made them completely flat it would be near impossible to pick them up with our robot's gripper when lying flat in a tray, which is very unrealistic, since in reality it is enough to press down almost anywhere on the bag and pinch it to grasp. If we modeled them as crumpled up, it would be too easy to grasp them and the policy would again not learn the pressing down and pinching behavior required to get a good grip on a flattened out bag. Preliminary experiments without deformable objects showed that the policy, when facing a chips bag in simulation or in the real world would just immediately give up and terminate, ``knowing'' that it wouldn't be able to grasp them.

\begin{figure}[ht!]
    \centering
    \includegraphics[width=1.0\linewidth]{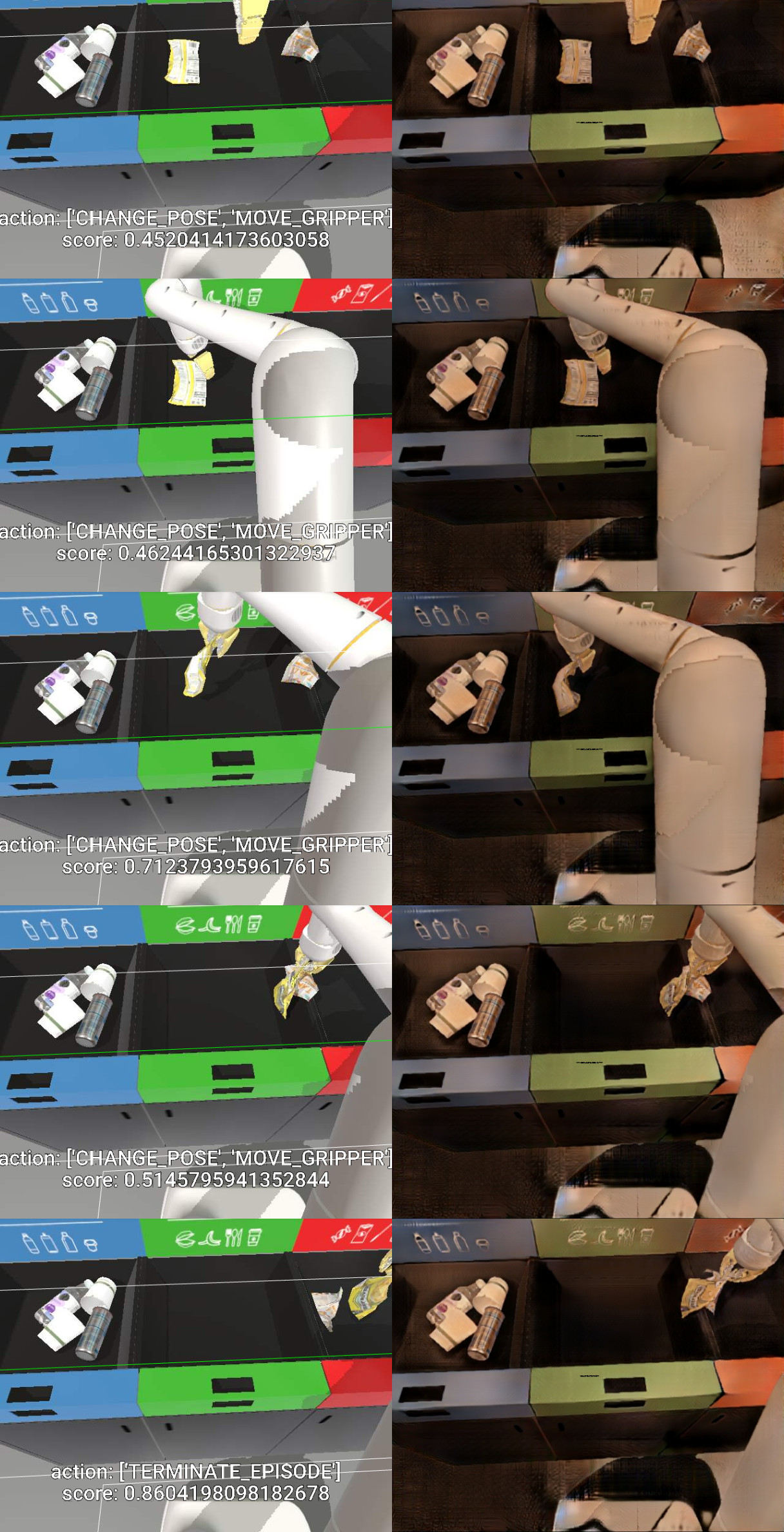}
    \caption{Example of manipulating a deformable object (a chips bag) in simulation. Images on the left show the original rendering from our simulation, on the right are the corresponding adapted images using RetinaGAN~\cite{ho2021retinagan}. The text on the left images shows the type of action chosen in the current step and the corresponding Q-value.}
    \label{fig:retinagan_and_deformable_examples}
\end{figure}

Figure~\ref{fig:retinagan_and_deformable_examples} shows an animation of the robot interacting with deformable objects in our simulation.

\subsection{Benchmark scenes taken from the ``wild''}

As described in the main paper, we create a set of scenarios representing actual situations our robots encountered at the deployment site (in the ``wild'') and use them for representative but repeatable experimentation in our \rc. Fig.~\ref{fig:waste_scenarios} shows images of the 9 in-distribution scenarios, Table~\ref{table:benchmark_scenes} presents a description of the distribution of objects in the individual bins.

\subsection{Object classes}

We define a list of 12 classes of objects we encountered in the waste stations in the wild and used to define our benchmark scenes, along with a mapping of those object classes to waste target bin categories as follows:
\begin{itemize}
   \item Recycle: ``can'', ``bottle'', ``drink carton'', ``yogurt cup''
   \item Compost: ``cup'', ``paper cup'', ``clear cup'', ``disposable bowl'', ``disposable plate'', ``paper container'', ``napkin''
   \item Landfill: ``bag/wrapper'', ``face mask''
\end{itemize}

When continuing to run the policy at the deployment site, whenever we would see objects that don't belong to any of those categories, we would define new object classes for our annotation and metrics computation process. The list of those additional classes includes ``non-disposable bowl'', ``non-disposable plate'',
``book/paper'', ``non-disposable drinkware'', ``food scraps'', and ``packaging''. The frequency at which those objects appeared in the ``wild'' during the time of our experimentation did not warrant including them in our benchmark scenes.

\newcommand{\misplaced}[1]{\mbox{\textcolor{red}{#1}}}

\begin{figure*}
\begin{tabularx}{\textwidth}{| c | >{\centering\arraybackslash}X | >{\centering\arraybackslash}X | >{\centering\arraybackslash}X|}
\hline
scene & recycle & compost & landfill \\
\hline\hline
1 
& \textcolor{red}{1 clear cup} 
& \mbox{1 cup}, \mbox{2 disposable bowls}, \mbox{2 napkins}, \misplaced{1 drink carton} 
&  \\ \hline
2 
& \mbox{2 bottles}, \mbox{6 cans} 
& \mbox{1 cup}, \mbox{2 napkins}
& \misplaced{2 paper containers}, \misplaced{1 cup}, \misplaced{4 napkins} \\ \hline
3 
& \mbox{1 can}, \mbox{1 bottle}, \misplaced{2 paper cup} 
& \misplaced{2 bag/wrappers} 
& \\ \hline
4 
& \mbox{1 bottle}, \misplaced{1 clear cup}, \misplaced{1 paper  cup}
& \mbox{1 paper cup}, \mbox{2 disposable bowls}, \mbox{1 napkin}, \misplaced{2 bag/wrappers} 
& \mbox{1 bag/wrapper}, \misplaced{1 yoghurt}, \misplaced{1 disposable plate} \\ \hline
5 
& \mbox{1 bottle}
& \mbox{1 paper cup}, \mbox{1 bottle}
& \misplaced{2 clear cups}, \misplaced{1 disposable bowl} \\ \hline
6  
& \mbox{2 yogurt cups}, \misplaced{1 clear cup}, \misplaced{1 bag/wrapper}, \misplaced{2 napkins} 
& \mbox{2 clear cups}, \mbox{4 paper cups}, \mbox{4 napkins}, \misplaced{1 bag/wrapper} 
& \misplaced{2 bottles}, \misplaced{2 napkins} \\ \hline
7 
& \misplaced{1 bag/wrapper} 
& \mbox{2 paper cups}, \mbox{1 clear cup}, \mbox{1 paper bag}
& \misplaced{1 yogurt cup}, \misplaced{2 napkins} \\ \hline
8 
& \mbox{5 cans}, \misplaced{1 clear cup}
& \mbox{1 disposable bowl}, \mbox{3 paper cups}, \mbox{1 clear cup}, \mbox{4 napkins}, \misplaced{1 can}, \misplaced{1 bag/wrapper}
& \misplaced{1 paper container}, \misplaced{1 disposable bowl}, \misplaced{2 napkins} \\ \hline
9
& \mbox{2 bottles}, \mbox{3 cans}, \misplaced{2 bag/wrappers}
& \mbox{1 paper container}, \mbox{1 paper cup}, \mbox{1 clear cup}, \mbox{4 disposable bowls}, \mbox{3 napkins}, \misplaced{1 yogurt cup}
& \\ \hline\hline
held out 1
& \mbox{2 bottles}, \mbox{8 cans}
& \mbox{3 paper cups}, \mbox{1 compostable bowl}, \mbox{1 napkin}
& \misplaced{1 banana peel}, \misplaced{1 clear cup}, \misplaced{4 napkins}, \misplaced{2 containers} \\ \hline
held out 2
& 1 bottle
& \misplaced{1 bottle} 
& \misplaced{2 clear cups}, \misplaced{1 compostable bowl}, 1 packaging \\ \hline
held out 3 
& \misplaced{1 bag/wrapper}
& \mbox{2 paper cups}, \mbox{1 clear cup}, \mbox{1 napkin}, \misplaced{1 face mask}
& \misplaced{2 napkins}, \misplaced{1 yogurt cup} \\ \hline 
\end{tabularx}
\caption{Description of the benchmark scenes used for evaluations, shown in Figure~\ref{fig:waste_scenarios}. A scene is defined by the initial quantity of objects of each class in each of the 3 waste bins. Misplaced objects are marked red. \label{table:benchmark_scenes}}
\end{figure*}

\subsection{Baseline performance of improved script}

After bootstrapping our flywheel with the script (Alg.~\ref{alg:scripted_expl}, line~\ref{alg:scripted_expl:script}), we further invested engineering time to extend the initial script with various strategies specific to some objects that the initial version struggled with. For example, we had to modify the approach direction and gripper opening to grasp large lunch boxes more reliably. We, however, did not add data generated with the more specialized script to our flywheel, but instead let our policies continue to improve through trial-and-error. The goal of this experiment was to test the hypothesis that performance gained through added engineering time can be met and surpassed with autonomous data-collection. We evaluate the extended script on the 9 waste scenarios we used for our evaluations in section~\ref{sec:experiments}. Overall, the improved script sorted $71\%$ of the objects which is higher than our initial policy trained with only sim data. However, our learned policy further improved as we spun the flywheel without added engineering time and eventually reached a performance of $84\%$.
Although, it remains unclear how the script compares to a learned policy at the limit of increasingly more engineering time investment versus data collection, we observe that our data-driven approach outperforms a fair amount of engineering effort by a margin of $13$ percentage points.

\end{document}